\documentclass[journal,twoside,web]{ieeecolor}
\usepackage{generic}
\usepackage{cite}
\usepackage{amsmath,amssymb,amsfonts}
\usepackage{graphicx}
\usepackage{algorithm,algorithmic}
\usepackage{hyperref}
\hypersetup{hidelinks=true}
\usepackage{textcomp}

\usepackage{booktabs}
\usepackage{multirow}
\usepackage{pifont}

\usepackage{bibunits}
\defaultbibliographystyle{IEEEtran}

\begin{document}
\begin{bibunit}
\nocite{IEEEexample:BSTcontrol}

\title{Curvature-Informed Potential Energy Surface for Protein-Ligand Binding Affinity Prediction}

\author{Peng-Fei Sun, Chuan-Xian Ren$^*$, Hong Yan
\thanks{This work is supported in part by National Key R\&D Program of China (2024YFA1011900), National Natural Science Foundation of China (62376291), Guangdong Basic and Applied Basic Research Foundation (2023B1515020004), Science and Technology Program of Guangzhou (2024A04J6413), and in part by the Hong Kong Innovation and Technology Commission (ITC) (InnoHK Project CIMDA) and the Institute of Digital Medicine of City University of Hong Kong (Project 9229503). \textit{(Corresponding author: Chuan-Xian Ren.)}}
\thanks{P.F. Sun and C.X. Ren are with the School of Mathematics, Sun Yat-Sen University, Guangzhou, Guangdong 510275, China (e-mail: rchuanx@mail.sysu.edu.cn). H. Yan is with the Department of Electrical Engineering, City University of Hong Kong, Hong Kong.}
\thanks{The code are available at https://github.com/Peng-Fei-Sun/CPES.}
}

\maketitle
\begin{abstract}
Accurate prediction of protein-ligand binding affinity is essential for structure-based drug discovery. Recent geometric deep learning methods have achieved promising performance by representing protein-ligand complexes as three-dimensional graphs. However, most existing approaches mainly rely on static interaction geometry from a single bound conformation, while neglecting molecular flexibility and binding-induced conformational changes. To address this limitation, we propose a curvature-informed potential energy surface (CPES) graph neural network for protein-ligand binding affinity prediction, which incorporates physics-informed curvature representations to model conformational flexibility. CPES first derives curvature spectral descriptors from the Hessian of the potential energy surface evaluated at equilibrium configurations, whose eigenvalues define the local principal curvatures of the potential energy surface. It then uses spectral cross-attention to compare the unbound ligand and protein with the bound complex, thereby capturing binding-induced changes in conformational dynamics. In parallel, hierarchical protein-ligand interaction representations are learned from static structural features through geometry-aware message passing, soft clustering, and bidirectional cross-attention. Finally, CPES fuses the curvature-informed dynamic representations with static interaction representations for affinity regression. Extensive evaluations on multiple benchmark datasets demonstrate that CPES achieves improved predictive performance and offers physical interpretability.
\end{abstract}


\section{Introduction}

\begin{figure*}[t]
    \centering
    \includegraphics[width=1\linewidth]{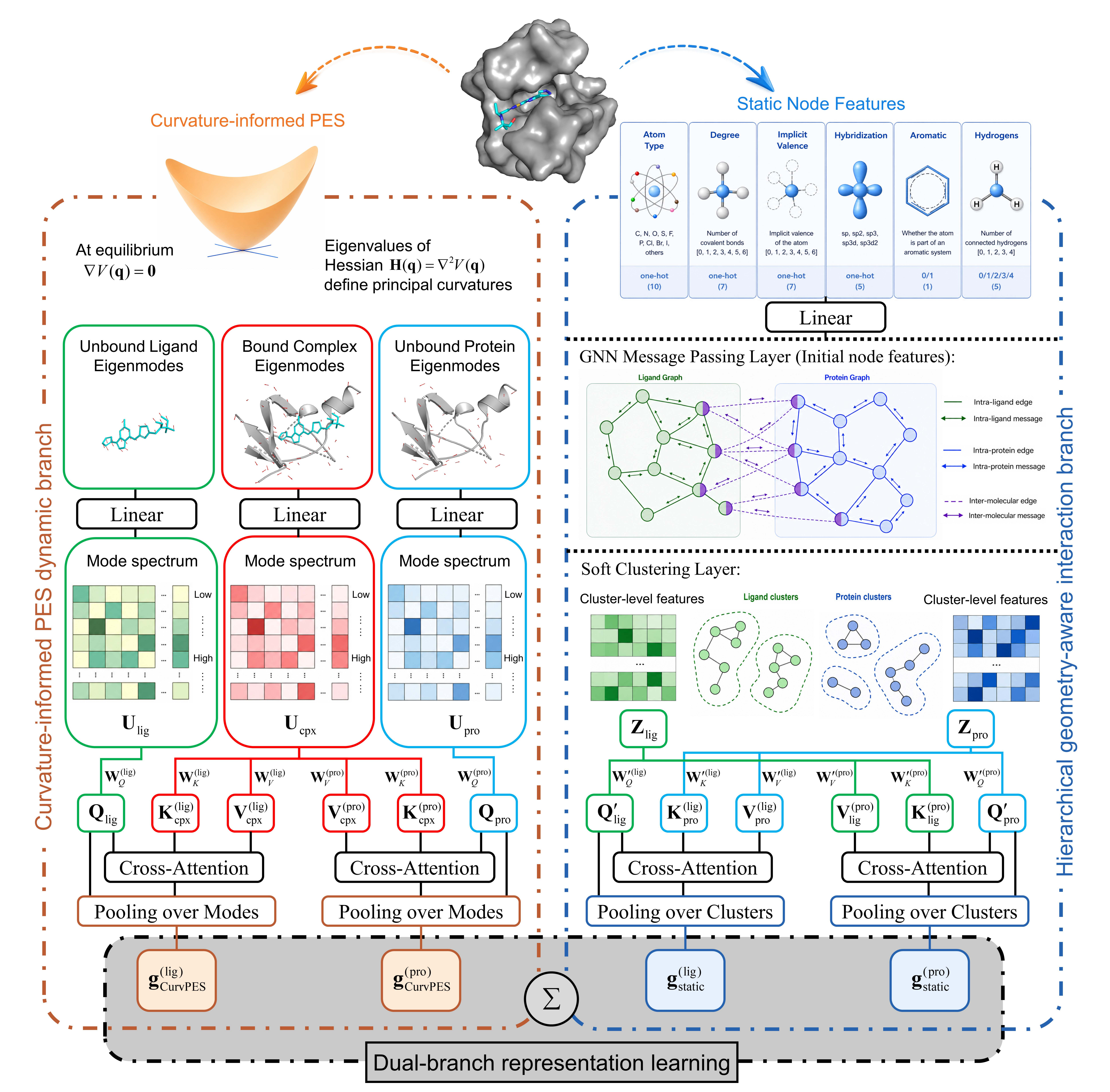}
    \caption{Overview of the CPES framework. CPES integrates curvature-informed dynamic modeling with geometry-aware structural learning. Eigenmodes characterize conformational flexibility and collective motions (small red arrows indicate molecular motion trends). Cross-attention is applied in both dynamic and static branches to capture binding-related interactions. By combining dynamic and structural information, the framework provides a physically meaningful representation for protein-ligand binding affinity prediction.}
    \label{fig:framework}
\end{figure*}

\IEEEPARstart{A}{ccurate} estimation of protein-ligand binding affinity, which reflects the strength of interaction between a biomolecular target and a small-molecule ligand, constitutes a cornerstone of structure-based drug discovery \cite{NEID5,NEID6,NEID7,NEID9}. This quantity underlies key stages of the discovery pipeline, including compound prioritization \cite{NEID10}, lead refinement \cite{NEID11}, and rational drug design \cite{NEID12}, and is commonly quantified using dissociation or inhibition constants such as $K_d$ and $K_i$. While experimental techniques offer reliable affinity measurements, their high cost, limited throughput, and substantial time requirements hinder large-scale exploration of chemical space. Thus, computational modeling has become an essential module of modern drug discovery, enabling efficient screening and ranking of candidate compounds \cite{NEID13}. The increasing availability of experimentally resolved protein-ligand complex structures, together with curated affinity annotations, has further stimulated the adoption of machine learning approaches for data-driven affinity prediction \cite{NEID15,NEID16}. Despite these advances, accurately capturing protein-ligand binding affinity remains a challenging task, owing to the intricate and context-dependent nature of molecular interactions, which continues to limit the effectiveness of existing learning-based models \cite{NEID17}. 

Recent advances in machine learning, particularly geometric deep learning, have substantially improved protein-ligand binding affinity prediction \cite{NEID4,NEID8,NEID14}. Existing methods can be broadly categorized into interaction-free and interaction-based approaches \cite{NEID4}. Interaction-free methods \cite{NEID18,NEID19,NEID20} represent ligands using fingerprints, SMILES, or 2D graphs and model proteins as sequences, intentionally omitting explicit atomic interactions for simplicity. In contrast, interaction-based models \cite{NEID21,NEID22,NEID23,NEID24,NEID25,NEID26,NEID27,NEID28,NEID29,NEID1,NEID2} represent protein-ligand complexes as three-dimensional graphs or grids and explicitly model atomic interactions and spatial relationships. By introducing geometric inductive biases that are more closely aligned with physical binding mechanisms, these interaction-based approaches have demonstrated superior predictive performance compared with sequence-based and handcrafted descriptor methods, and have become a dominant paradigm for protein-ligand interaction modeling, particularly when implemented using interaction graph neural networks. A more detailed discussion of related works and additional references is provided in  Supplement~\ref{appendix:Related Works}.

Despite their success, most existing graph-based approaches, including state-of-the-art interaction-based methods, still implicitly treat protein-ligand complexes as rigid entities and rely primarily on static geometric features extracted from a single bound conformation. Such assumptions overlook the inherently dynamic nature of molecular binding, which occurs over ensembles of conformations rather than fixed structures. Moreover, most methods do not explicitly account for the discrepancy in conformational dynamics between the apo (unbound) and holo (bound) states of protein-ligand complexes. From a biophysical perspective, protein-ligand binding is governed by the underlying potential energy surface, which governs the relative stability and conformational flexibility of apo and holo states. Although explicitly computing the full potential energy surface is computationally infeasible at scale, its local structure encodes essential information about molecular stiffness and flexibility that can be captured through curvature.

Motivated by this observation, we incorporate descriptors inspired by such local conformational energy variations as curvature-informed priors within geometric representation learning and propose a curvature-informed potential energy surface (CPES) graph neural network for protein-ligand binding affinity prediction. By integrating these priors into a hierarchical protein-ligand graph framework, CPES captures molecular flexibility and dynamic compatibility alongside interaction geometry, capturing both protein-ligand interactions and binding-induced conformational changes. In summary, the main contributions of our work are as follows.
\begin{itemize}
\item A physical energy curvature-informed inductive bias is introduced for protein-ligand binding affinity prediction. We formulate the binding affinity prediction from a physics-informed perspective and argue that incorporating energy-related geometric information reflecting physical and biological binding mechanisms is critical for generalization and interpretability.
\item A curvature-informed potential energy surface (CPES) graph neural network is proposed for geometric representation learning. CPES integrates curvature-derived descriptors, defined as the eigenvalues of the Hessian of the potential energy surface at equilibrium (i.e., the principal curvatures), enabling the model to capture molecular flexibility and dynamic compatibility beyond static interaction geometry.
\item CPES models the discrepancy between apo (unbound) and holo (bound) conformational dynamics via cross-attention over curvature-informed representations, capturing binding-induced conformational changes and achieving improved predictive performance and more biologically meaningful modeling across diverse benchmarks.
\end{itemize}

\section{Curvature of the Potential Energy Surface}

The stability and flexibility of protein-ligand systems in configuration space are fundamentally governed by the local curvature of the potential energy surface near equilibrium. The potential energy function defines how the system's energy varies with molecular conformation, and under physiological conditions, conformational fluctuations are typically confined to small deviations around an equilibrium or metastable state. Consequently, the local shape of the potential energy surface in this region provides a natural physical description of conformational behavior.

Consider a system composed of $N$ interaction sites, whose configuration is described by the Cartesian coordinate vector
\begin{equation*}
   \mathbf{q}\in {{\mathbb{R}}^{3N}}, 
\end{equation*}
where each triplet corresponds to the three-dimensional position of a site. Let $\mathbf{q}_0$ denote an equilibrium conformation. In the vicinity of $\mathbf{q}_0$, the potential energy function $V(\mathbf{q})$ can be expanded as
\begin{equation*}
\begin{split}
V(\mathbf{q}) =\;& V(\mathbf{q}_0) + \nabla V(\mathbf{q}_0)^\top (\mathbf{q} - \mathbf{q}_0) \\
&+ \frac{1}{2} (\mathbf{q} - \mathbf{q}_0)^\top \mathbf{H} (\mathbf{q} - \mathbf{q}_0) + o\!\left(\|\mathbf{q} - \mathbf{q}_0\|^2\right),
\end{split}
\end{equation*}
where $\nabla V(\mathbf{q}_0)$ is the first derivative of the potential energy, and $\mathbf{H}\in {{\mathbb{R}}^{3N\times 3N}}$ is the Hessian matrix with elements
\begin{equation*}
    {{H}_{ij}}={{\left. \frac{{{\partial }^{2}}V}{\partial {{q}_{i}}\partial {{q}_{j}}} \right|}_{\mathbf{q}={{\mathbf{q}}_{0}}}}. 
\end{equation*}
At equilibrium, the net force acting on the system vanishes, implying $\nabla V({{\mathbf{q}}_{0}})=\mathbf{0}$. Under this condition, the local shape of the potential energy surface is fully determined by the second-order term, and the Hessian provides a complete description of local curvature.

Because the Hessian is a real symmetric matrix constructed from second derivatives of the potential energy, it admits an orthogonal spectral decomposition
\begin{equation*}
\mathbf{H} = \mathbf{\Phi} \,\mathbf{\Lambda}\, \mathbf{\Phi}^\top,
\qquad
\mathbf{\Lambda} = \operatorname{diag}(\lambda_1, \lambda_2, \ldots, \lambda_{3N}),
\end{equation*}
where the columns of $\mathbf{\Phi}$ define orthonormal mode directions. In this eigenbasis, the second-order variation of the potential energy is fully decoupled, such that each eigenvalue $\lambda_k$ directly quantifies the local curvature of the energy surface along orthogonal directions, which can be interpreted as principal curvatures in a generalized high-dimensional setting. At a local minimum of the potential energy, all non-zero eigenvalues are positive, each eigenvalue corresponds to the second-order variation of the potential energy along its associated eigenvector and therefore quantifies the local curvature in that direction. Eigenvalues that are exactly zero correspond to motions that do not change the internal potential energy. For three-dimensional molecular systems, the Hessian therefore typically exhibits at least six zero eigenvalues, associated with rigid-body translations and rotations.

In summary, the spectral structure of the Hessian provides a compact and physically interpretable characterization of the local curvature of the potential energy surface. Low-curvature modes dominate conformational changes because displacements along these directions are energetically inexpensive, allowing large-scale collective rearrangements, whereas high-curvature modes strongly resist deformation. This curvature-based perspective motivates the use of low-curvature spectral information to describe global flexibility and collective conformational behavior in subsequent modeling.

\textbf{SE(3) Invariance of the Hessian Eigenvalues.} To ensure that curvature-based descriptors are independent of the global coordinate frame, we note that the eigenvalue spectrum of the Hessian is invariant under rigid-body transformations in SE(3). Because the potential energy depends only on internal geometric relationships, it is invariant under global rigid-body transformations. In particular, global translations do not alter the Hessian matrix, as second derivatives of the potential energy are unaffected by constant shifts of the coordinates. Global rotations correspond to orthogonal changes of basis in the 3N-dimensional configuration space, under which the Hessian undergoes an orthogonal similarity transformation. Such transformations preserve the eigenvalue spectrum, implying that Hessian eigenvalues are invariant under SE(3) transformations, while the corresponding eigenvectors rotate with the coordinate frame. This property can be formally stated as the following proposition.

\textbf{Proposition (SE(3) equivariance of the Hessian).} A potential energy function and its Hessian matrix are given as follows
\begin{equation}
\begin{aligned}
V(\mathbf{q}) \;:&\; \mathbb{R}^{3N} \to \mathbb{R}, \\
\mathbf{H}(\mathbf{q}) \;=&\; \nabla^2 V(\mathbf{q}) 
= \frac{\partial^2 V(\mathbf{q})}{\partial \mathbf{q} \, \partial \mathbf{q}^{\top}} 
\in \mathbb{R}^{3N \times 3N}.
\end{aligned}
\end{equation}
The potential energy function depends only on the internal geometric relations of the system and is invariant under global rigid-body transformations. Specifically, for any rotation and any translation $\mathbf{R} \in \mathrm{SO}(3),\; \mathbf{t} \in \mathbb{R}^3,\; (\mathbf{R},\mathbf{t}) \in \mathrm{SE}(3)$ define
\begin{equation*}
\begin{aligned}
\mathbf{Q} &= \operatorname{diag}(\underbrace{\mathbf{R}, \ldots, \mathbf{R}}_{N}) \in \mathbb{R}^{3N \times 3N}, \\
\mathbf{c} &= (\underbrace{\mathbf{t}, \ldots, \mathbf{t}}_{N}) \in \mathbb{R}^{3N}.
\end{aligned}
\end{equation*}
The SE(3)-transformed configuration is given by
\begin{equation*}
    \mathbf{q}' = \mathbf{Q}\,\mathbf{q} + \mathbf{c}.
\end{equation*}
Then, under an SE(3) transformation, the potential energy function $V'(\mathbf{q}')$ and its Hessian matrix $\mathbf{H}'(\mathbf{q}')$ satisfy
\begin{equation}
    V'(\mathbf{q}') \equiv V(\mathbf{q}), \qquad 
    \mathbf{H}'(\mathbf{q}') = \mathbf{Q}\,\mathbf{H}(\mathbf{q})\,\mathbf{Q}^\top.
\end{equation}
The equality $V'(\mathbf{q}') \equiv V(\mathbf{q})$ follows directly from the physical invariance of the potential energy under global rigid-body transformations. The corresponding relation for the Hessian matrix $\mathbf{H}'(\mathbf{q}') = \mathbf{Q}\,\mathbf{H}(\mathbf{q})\,\mathbf{Q}^\top$ is a mathematical consequence of this invariance and can be rigorously derived via the chain rule under coordinate reparameterization (detailed derivation is provided in  Supplement~\ref{appendix:proof}). 

Since $\mathbf{Q}$ is an orthogonal matrix satisfying $\mathbf{Q}^\top \mathbf{Q} = \mathbf{I}$, the relation between the Hessian matrices before and after the SE(3) transformation constitutes an orthogonal similarity transformation. Consequently, the eigenvalues of the Hessian are invariant, while the eigenvectors transform covariantly with the coordinate system, i.e.,
\begin{equation}
    \lambda_k' = \lambda_k, \qquad 
    \boldsymbol{\varphi}_k' = \mathbf{Q}\,\boldsymbol{\varphi}_k, \quad k = 1, \ldots, 3N - 6,
\end{equation}
where $k$ indexes the normal modes of the Hessian matrix excluding the rigid-body zero modes.

\section{Methods}

\subsection{Descriptors of Potential Energy Function Curvature}

To obtain a computable approximation of the second-order geometry of the potential energy surface, we adopt the anisotropic network model (ANM) to represent the potential energy function in the vicinity of a reference configuration. The core assumptions of ANM are encoded in the functional form of the potential energy: near the reference configuration the potential energy depends solely on inter-node distance fluctuations and is approximated by an anisotropic harmonic form
\begin{equation*}
    V(\mathbf{q}) = \frac{1}{2} \sum_{(i,j)\in \mathcal{E}}
    \left( \lVert \mathbf{r}_{ij} \rVert - \lVert \mathbf{r}_{ij}^{0} \rVert \right)^2.
\end{equation*}
Here, ${{\mathbf{r}}_{i}}\in {{\mathbb{R}}^{3}}$ denotes the three-dimensional coordinate vector of node $i$, ${{\mathbf{r}}_{ij}}={{\mathbf{r}}_{j}}-{{\mathbf{r}}_{i}}$ denotes the relative displacement vector between nodes $i$ and $j$ in the current configuration, $\mathbf{r}_{ij}^{0}=\mathbf{r}_{j}^{0}-\mathbf{r}_{i}^{0}$ denotes the corresponding relative displacement vector in the reference configuration, and $\mathcal{E}$ is the set of undirected interacting node pairs. This formulation implicitly assumes small-amplitude fluctuations compatible with the harmonic approximation and, critically, that each interaction constrains motion only along its reference direction while transverse displacements do not contribute to the restoring force. 

For each interacting pair $i$ and $j$, a unit direction vector is defined as
\begin{equation*}
    \mathbf{e}_{ij} = \frac{\mathbf{r}_{ij}^{0}}{\lVert \mathbf{r}_{ij}^{0} \rVert},
\end{equation*}
which specifies the geometric orientation of the interaction in three-dimensional space. Since the potential energy is sensitive only to distance changes along this direction, the second-order contribution of the interaction naturally takes the form of a projection onto $\mathbf{e}_{ij}$, leading to the local stiffness matrix which characterizes the resistance of a local molecular region to small structural perturbations, i.e., 
\begin{equation*}
    \mathbf{K}_{ij} = \mathbf{e}_{ij}\,\mathbf{e}_{ij}^\top \in \mathbb{R}^{3 \times 3}.
\end{equation*}
This rank-one matrix penalizes only displacement components aligned with the interaction direction while leaving transverse motions unconstrained, thereby providing a direct mathematical realization of the anisotropic assumption. The global Hessian is assembled from these $3\times3$ stiffness blocks $\mathbf{K}_{ij}$ in an edge-wise manner. For each undirected interacting node pairs $(i,j)\in \mathcal{E}$, the following block updates are applied
\begin{equation}
\label{eq:K_update}
\begin{aligned}
\mathbf{H}_{ii} \mathrel{+}= \mathbf{K}_{ij}, \quad
\mathbf{H}_{jj} \mathrel{+}= \mathbf{K}_{ij}, \\
\mathbf{H}_{ij} \mathrel{-}= \mathbf{K}_{ij}, \quad
\mathbf{H}_{ji} \mathrel{-}= \mathbf{K}_{ij},
\end{aligned}
\end{equation}
where $\mathbf{H}_{ij} \in \mathbb{R}^{3 \times 3}$ denotes the $(i,j)$-th $3\times3$ sub-block of the Hessian matrix. This construction corresponds to expanding the quadratic energy contribution of each interaction and, by construction, ensures symmetry, positive semidefiniteness, and invariance of the potential energy under rigid-body translations. From this edge-wise assembly, the characteristic block structure of the Hessian becomes evident: off-diagonal blocks encode pairwise couplings, while diagonal blocks accumulate contributions from all interactions incident to each node. Accordingly, the assembled Hessian admits the following piecewise representation
\begin{equation}
\label{eq:K_assembled}
\mathbf{H}_{ij} =
\begin{cases}
-\mathbf{K}_{ij}, & i \ne j,\ (i,j) \in \mathcal{E}, \\
\displaystyle \sum_{k:(i,k)\in \mathcal{E}} \mathbf{K}_{ik}, & i = j, \\
\mathbf{0}, & \text{otherwise}.
\end{cases}
\end{equation}

In the construction of Eqs.~\eqref{eq:K_update} and \eqref{eq:K_assembled}, the Hessian matrix $\mathbf{H}$ exhibits a typical graph-Laplacian-like block structure. The diagonal blocks are given by the summation of all neighboring edge contributions, whereas the off-diagonal blocks are given by the negative contribution of the corresponding edge. This form generalizes the classical scalar graph Laplacian $\mathbf{L}=\mathbf{D}-\mathbf{A}$ (where $\mathbf{D}$ denotes the degree matrix and $\mathbf{A}$ denotes the adjacency matrix) to a vector-valued graph Laplacian, where each edge $(i,j)$ is assigned a $3\times 3$ anisotropic stiffness matrix $\mathbf{K}_{ij}$ rather than a single scalar weight. Therefore, $\mathbf{H}$ can be viewed as a second-order elliptic discrete operator defined on the space of node displacement vectors, and its quadratic form directly corresponds to the elastic potential energy of the system around the reference conformation. This perspective unifies the stiffness matrix in mechanics within the framework of spectral graph theory, providing a natural algebraic foundation for the subsequent eigendecomposition and low-frequency spectral feature extraction.

Following Hessian construction, symmetric eigendecomposition is performed 
\begin{equation*}
    \mathbf{H}\,\boldsymbol{\varphi}_k = \lambda_k \boldsymbol{\varphi}_k,
\end{equation*}
where the eigenvalues $\lambda_k$ quantify the local curvature of the potential energy surface along orthogonal normal modes $\boldsymbol{\varphi}_k$. To remove rigid-body and near-zero modes, a relative cutoff is applied, retaining only modes satisfying
\begin{equation*}
    \lambda_k > \varepsilon, \qquad
    \varepsilon = \varepsilon_{\mathrm{rel}} \max_{k} \lvert \lambda_k \rvert.
\end{equation*}
Among the remaining modes (we apply $\varepsilon_{\mathrm{rel}} = 1e$-6), the smallest $K_{\mathrm{u}}$ eigenvalues are selected as global descriptors of potential energy function curvature. Under the harmonic approximation, the eigendecomposition of the Hessian reveals that low-frequency eigenvalues correspond to the flattest directions of the potential energy surface and thus dominate global flexibility and conformational accessibility. A logarithmic transform with standard numerical stabilization yields the graph-level curvature feature
\begin{equation*}
\mathbf{u} = \left[ \ln \lambda_1,\ \ln \lambda_2,\ \ldots,\ \ln \lambda_{K_{\mathrm{u}}} \right].
\end{equation*}

The above procedure is applied consistently across different conformational states. The only difference lies in how the node set and the undirected interaction edge set $\mathcal{E}$ are constructed. We build descriptors for the unbound ligand, the unbound protein, and the bound complex, yielding $\mathbf{u}_{\text{lig}},\ \mathbf{u}_{\text{pro}},\ \mathbf{u}_{\text{cpx}}$ respectively.

\subsubsection{Unbound Ligand}
The node set ${{\mathcal{V}}_{\text{lig}}}$ consists of ligand atoms, and the interaction set ${{\mathcal{E}}_{\text{lig}}}$ contains intra-ligand covalent bonds. The ligand Hessian and its spectral feature are
\begin{equation}
\mathbf{H}_{\text{lig}} = \mathbf{H}(\mathcal{V}_{\text{lig}}, \mathcal{E}_{\text{lig}}), \quad
\mathbf{u}_{\text{lig}} = \left[ \ln \lambda_{1}^{\text{lig}}, \ldots, \ln \lambda_{K_{\mathrm{ul}}}^{\text{lig}} \right].
\end{equation}
This feature characterizes the intrinsic curvature spectrum and low-frequency flexibility of the ligand in the unbound state.

\subsubsection{Unbound Protein}
The node set ${{\mathcal{V}}_{\text{pro}}}$ consists of protein residues represented by $C_\alpha$ atoms. The interaction set ${{\mathcal{E}}_{\text{pro}}}$ is defined on the residue graph by projecting atomic covalent connectivity onto residue-residue interactions. Accordingly,
\begin{equation}
\mathbf{H}_{\text{pro}} = \mathbf{H}(\mathcal{V}_{\text{pro}}, \mathcal{E}_{\text{pro}}), \quad
\mathbf{u}_{\text{pro}} = \left[ \ln \lambda_{1}^{\text{pro}}, \ldots, \ln \lambda_{K_{\mathrm{up}}}^{\text{pro}} \right].
\end{equation}
This feature captures the intrinsic curvature spectrum of the protein pocket in the unbound state.

\subsubsection{Bound Complex}
For the bound complex, ligand and protein nodes are unified as ${{\mathcal{V}}_{\text{cpx}}}={{\mathcal{V}}_{\text{lig}}}\cup {{\mathcal{V}}_{\text{pro}}}$, and the interaction set includes intra-ligand interactions ${{\mathcal{E}}_{\text{lig}}}$, intra-protein interactions ${{\mathcal{E}}_{\text{pro}}}$, and inter-molecular interactions ${{\mathcal{E}}_{\text{inter}}}$. The inter-molecular set is defined by a distance cutoff hyperparameter ${{d}_{\text{cutoff}}}$, i.e.,
\begin{equation}
\label{eq:dcutoff}
\mathcal{E}_{\text{inter}} =
\left\{ (i,j) \;\middle|\;
i \in \mathcal{V}_{\text{lig}},\;
j \in \mathcal{V}_{\text{pro}},\;
\lVert \mathbf{r}_i - \mathbf{r}_j \rVert < d_{\text{cutoff}}
\right\}.
\end{equation}
Thus ${{\mathcal{E}}_{\text{cpx}}}={{\mathcal{E}}_{\text{lig}}}\cup {{\mathcal{E}}_{\text{pro}}}\cup {{\mathcal{E}}_{\text{inter}}}$ and the bound complex Hessian is constructed
\begin{equation}
\mathbf{H}_{\text{cpx}} = \mathbf{H}(\mathcal{V}_{\text{cpx}}, \mathcal{E}_{\text{cpx}}), \quad
\mathbf{u}_{\text{cpx}} = \left[ \ln \lambda_{1}^{\text{cpx}}, \ldots, \ln \lambda_{K_{\mathrm{uc}}}^{\text{cpx}} \right].
\end{equation}
Since ${{\mathbf{H}}_{\text{cpx}}}$ explicitly introduces ligand-protein couplings, the graph-level feature ${{\mathbf{u}}_{\text{cpx}}}$ represents the curvature spectrum of the bound-state ligand-protein complex rather than a trivial concatenation of unbound spectra of them.

To ensure a consistent dimensionality of the spectral descriptors, we apply a padding strategy when the number of non-zero eigenvalues is smaller than the predefined spectral size (i.e., $K_{\mathrm{ul}}$, $K_{\mathrm{up}}$, or $K_{\mathrm{uc}}$). Specifically, if the available non-zero eigenvalues are insufficient, the remaining entries are filled by repeating the last (i.e., largest-index) non-zero eigenvalue.

\subsection{Cross-Attention on Binding-Induced Conformational Changes}

To characterize binding-induced conformational changes, we introduce a spectral cross-attention mechanism that explicitly aligns and contrasts conformational responses among the unbound ligand, the unbound protein, and the bound ligand-protein complex. Unlike representations derived purely from static structures, this mechanism focuses on binding-induced changes in conformational response patterns, i.e., how the system redistributes its sensitivity to dynamic perturbations across conformational directions. Physically, such differences can be interpreted as changes in the curvature of the potential energy surface, and the resulting attention weights provide a data-driven indication of which conformational directions are most strongly reshaped upon binding.

Let $d$ denote the hidden dimension. We construct spectral representations for the ligand, protein, and complex as described above
\begin{equation*}
\mathbf{u}_{\text{lig}} \in \mathbb{R}^{K_{\mathrm{ul}}}, \quad
\mathbf{u}_{\text{pro}} \in \mathbb{R}^{K_{\mathrm{up}}}, \quad
\mathbf{u}_{\text{cpx}} \in \mathbb{R}^{K_{\mathrm{uc}}},
\end{equation*}
where $K_{\mathrm{ul}}$, $K_{\mathrm{up}}$, $K_{\mathrm{uc}}$ denote the numbers of retained spectral modes. Intuitively, each spectral component characterizes the stiffness or response magnitude along a particular conformational direction, facilitating quantitative comparison of binding-induced differences in conformational responses. Each spectral scalar is treated as a token and embedded into a $d$-dimensional feature space using a multi-layer perceptron (MLP), yielding
\begin{equation*}
\mathbf{U}_{\text{lig}} \in \mathbb{R}^{K_{\mathrm{ul}} \times d}, \quad
\mathbf{U}_{\text{pro}} \in \mathbb{R}^{K_{\mathrm{up}} \times d}, \quad
\mathbf{U}_{\text{cpx}} \in \mathbb{R}^{K_{\mathrm{uc}} \times d}.
\end{equation*}

To contrast conformational response patterns before binding (ligand or protein) and after binding (complex), we apply spectral cross-attention, where ligand or protein spectra serve as query and complex spectra serve as key and value. For $s\in\{\text{lig},\text{pro}\}$, the query, key, and value matrices are given by
\begin{equation*}
\begin{aligned}
\mathbf{Q}_s &= \mathbf{U}_s \mathbf{W}_Q^{(s)} \in \mathbb{R}^{K_{\mathrm{us}} \times d}, \\
\mathbf{K}_{\text{cpx}}^{(s)} &= \mathbf{U}_{\text{cpx}} \mathbf{W}_K^{(s)} \in \mathbb{R}^{K_{\mathrm{uc}} \times d}, \\
\mathbf{V}_{\text{cpx}}^{(s)} &= \mathbf{U}_{\text{cpx}} \mathbf{W}_V^{(s)} \in \mathbb{R}^{K_{\mathrm{uc}} \times d},
\end{aligned}
\end{equation*}
where $K_{\mathrm{us}} = K_{\mathrm{ul}} \text{ for } s = \text{lig}, \quad K_{\mathrm{us}} = K_{\mathrm{up}} \text{ for } s = \text{pro}$. $\mathbf W_Q^{(s)}, \mathbf W_K^{(s)}, \mathbf W_V^{(s)} \in \mathbb R^{d\times d}$ are learnable weight matrices. The spectral cross-attention is then computed as
\begin{equation*}
\begin{aligned}
\widetilde{\mathbf{U}}_{s \to \text{c}} &=
\operatorname{softmax}\!\left(
\frac{\mathbf{Q}_s\,\mathbf{K}_{\text{cpx}}^{(s)\top}}{\sqrt{d}}
\right)\,\mathbf{V}_{\text{cpx}}^{(s)}
\in \mathbb{R}^{K_{\mathrm{us}} \times d}.
\end{aligned}
\end{equation*}
This formulation uses the bound complex as a reference to match and reweight ligand or protein contributions across spectral modes, thereby emphasizing conformational directions that are most altered upon binding. In this way, the attention distribution implicitly captures structured differences in conformational responses and, consequently, in the curvature of the potential energy surface between the unbound components and the bound complex.

Finally, we aggregate the sequence-level responses into a graph-level vector and combine them with a query-side residual, resulting in a potential energy surface (PES) curvature-aware representation 
\begin{equation}
\label{eq:gCurvPES}
\mathbf{g}_{\text{CurvPES}}^{(s)} =
\operatorname{MLP}\!\left( \sum_{i=1}^{K_{\mathrm{us}}} \widetilde{\mathbf{U}}_{s \to \text{c}, i} \right)
+ \sum_{i=1}^{K_{\mathrm{us}}} \mathbf{U}_{s,i}
\in \mathbb{R}^{d}.
\end{equation}
This representation summarizes binding-induced changes in conformational response patterns and captures interpretable features related to variations in energy curvature.

\begin{table*}[t]
\centering
\caption{Performance Comparison of CPES and Baselines on the 2013 core set, 2016 core set, and the 2019 holdout set}
\label{table:Cross-Dataset}
\resizebox{\textwidth}{!}{
\begin{tabular}{llcccccc}
\toprule
\multirow{2}{*}{} & \multirow{2}{*}{Model} 
& \multicolumn{2}{c}{2013 core set} 
& \multicolumn{2}{c}{2016 core set} 
& \multicolumn{2}{c}{2019 holdout set} \\
\cmidrule(lr){3-4} \cmidrule(lr){5-6} \cmidrule(lr){7-8}
& & RMSE $\downarrow$ & Pearson $\uparrow$ 
& RMSE $\downarrow$ & Pearson $\uparrow$ 
& RMSE $\downarrow$ & Pearson $\uparrow$ \\
\midrule

\multirow{6}{*}{Interaction-free}
& DeepDTA \cite{NEID18} & 1.639 (0.026) & 0.718 (0.014) & 1.357 (0.015) & 0.785 (0.007) & 1.485 (0.023) & 0.586 (0.012) \\
& GraphDTA-GCN \cite{NEID19} & 1.749 (0.062) & 0.662 (0.032) & 1.513 (0.048) & 0.719 (0.023) & 1.763 (0.039) & 0.439 (0.021) \\
& GraphDTA-GAT \cite{NEID19} & 2.043 (0.029) & 0.476 (0.022) & 1.748 (0.019) & 0.594 (0.010) & 1.663 (0.027) & 0.432 (0.016) \\
& GraphDTA-GIN \cite{NEID19} & 1.691 (0.124) & 0.694 (0.059) & 1.470 (0.065) & 0.743 (0.027) & 1.676 (0.032) & 0.472 (0.021) \\
& GraphDTA-GAT-GCN \cite{NEID19} & 1.645 (0.085) & 0.711 (0.036) & 1.434 (0.064) & 0.754 (0.025) & 1.705 (0.075) & 0.474 (0.028) \\
& MGraphDTA \cite{NEID20} & 1.680 (0.093) & 0.696 (0.046) & 1.439 (0.047) & 0.753 (0.022) & 1.553 (0.028) & 0.538 (0.013) \\
\midrule

\multirow{12}{*}{Interaction-based}
& RF-Score \cite{NEID106} & 1.616 (0.010) & 0.776 (0.003) & 1.488 (0.001) & 0.785 (0.001) & 1.406 (0.001) & 0.618 (0.001) \\
& Pafnucy \cite{NEID15} & 1.517 (0.014) & 0.783 (0.005) & 1.450 (0.047) & 0.769 (0.019) & 1.438 (0.016) & 0.612 (0.014) \\
& OnionNet \cite{NEID108} & 1.583 (0.079) & 0.741 (0.037) & 1.399 (0.076) & 0.770 (0.027) & 1.510 (0.034) & 0.573 (0.014) \\
& PotentialNet \cite{NEID21} & 1.607 (0.027) & 0.773 (0.010) & 1.503 (0.033) & 0.772 (0.007) & 1.514 (0.028) & 0.564 (0.014) \\
& SchNet \cite{NEID22} & 1.570 (0.029) & 0.754 (0.030) & 1.390 (0.023) & 0.787 (0.016) & 1.522 (0.071) & 0.560 (0.028) \\
& GNN-DTI \cite{NEID23} & 1.533 (0.084) & 0.767 (0.040) & 1.384 (0.013) & 0.779 (0.008) & 1.446 (0.006) & 0.614 (0.007) \\
& IGN \cite{NEID24} & 1.428 (0.020) & 0.807 (0.001) & 1.269 (0.030) & 0.821 (0.013) & 1.410 (0.015) & 0.630 (0.008) \\
& EGNN \cite{NEID25} & 1.498 (0.025) & 0.782 (0.015) & 1.289 (0.021) & 0.816 (0.011) & 1.399 (0.013) & 0.628 (0.010) \\
& GIGN \cite{NEID1} & 1.380 (0.009) & 0.821 (0.003) & 1.190 (0.017) & 0.840 (0.007) & 1.393 (0.007) & 0.641 (0.006) \\
& MetalProGNet \cite{NEID109} & 1.494 (0.027) & 0.773 (0.012) & 1.309 (0.043) & 0.802 (0.013) & 1.448 (0.021) & 0.610 (0.009) \\
& SS-GNN \cite{NEID86} & 1.330 (0.011) & 0.830 (0.007) & 1.165 (0.011) & 0.846 (0.004) & 1.450 (0.006) & 0.633 (0.004) \\
& EHIGN \cite{NEID91} & 1.297 (0.026) & 0.841 (0.006) & 1.150 (0.022) & 0.854 (0.004) & 1.368 (0.021) & 0.667 (0.008) \\
\midrule

\multirow{6}{*}{\begin{tabular}[c]{@{}c@{}}Interaction-based\\(attention\\mechanism)\end{tabular}}
& AttentionSiteDTI \cite{NEID26} & 1.444 (0.037) & 0.792 (0.014) & 1.352 (0.022) & 0.784 (0.008) & 1.539 (0.015) & 0.563 (0.004) \\
& CAPLA \cite{NEID27} & 1.409 & 0.816 & 1.206 & 0.841 & - & - \\
& GAABind \cite{NEID28} & 1.488 & 0.772 & 1.297 & 0.803 & - & - \\
& DEAttentionDTA \cite{NEID29} & 1.470 & 0.800 & 1.266 & 0.827 & - & - \\
& CheapNet \cite{NEID2} & 1.262 (0.017) & 0.857 (0.004) & 1.107 (0.011) & 0.870 (0.002) & 1.343 (0.007) & 0.665 (0.003) \\
& CPES (ours) & \textbf{1.193 (0.010)} & \textbf{0.876 (0.006)} & \textbf{1.101 (0.009)} & \textbf{0.870 (0.003)} & \textbf{1.321 (0.011)} & \textbf{0.678 (0.007)} \\
\bottomrule
\end{tabular}
}
\end{table*}

\subsection{The Framework of CPES}

An overview of the CPES framework is illustrated in Fig.~\ref{fig:framework}. CPES follows a dual-branch architecture that integrates geometry-aware static interaction modeling with curvature-informed potential energy surface (PES) dynamics. Given a complex graph $\mathcal{G}=(\mathcal{V},\mathcal{E})$ the node set is defined as $\mathcal V=\mathcal V_{\text{lig}}\cup\mathcal V_{\text{pro}}$, and the edge set as $\mathcal E=\mathcal E_{\mathrm{intra}}\cup\mathcal E_{\mathrm{inter}}$ (where ${{\mathcal{E}}_{\mathrm{intra}}}={{\mathcal{E}}_{\text{lig}}}\cup {{\mathcal{E}}_{\text{pro}}}$), representing intra-molecular and inter-molecular interactions, respectively. Each atom is associated with an initial feature vector and a three-dimensional coordinate. These atom-level features are first mapped into a hidden representation space through a learnable embedding layer.

The static branch employs graph neural networks (GNNs) over both intra-molecular and inter-molecular edges of the complex graph to capture structural interaction information. Based on the atom-level representations, soft clustering is applied separately to the ligand and protein subgraphs to obtain hierarchical cluster-level features, denoted as $\mathbf Z_{\text{lig}}$ and $\mathbf Z_{\text{pro}}$. A bidirectional cross-attention module is then introduced between ligand and protein clusters to capture key structural interaction regions, yielding two graph-level static representations $\mathbf g_{\text{static}}^{(\text{lig})}$ and $\mathbf g_{\text{static}}^{(\text{pro})}$ (see  Supplement~\ref{appendix:framework_static} for detailed formulations and architectural descriptions).

In parallel, the curvature-informed PES branch uses the Hessian eigenspectrum to characterize conformational flexibility and collective dynamics beyond static interaction geometry. As defined in Eq.~\eqref{eq:gCurvPES}, this branch compares the unbound ligand and protein with the bound complex in the spectral space, producing $\mathbf g_{\text{CurvPES}}^{(\text{lig})}$ and $\mathbf g_{\text{CurvPES}}^{(\text{pro})}$ to describe curvature-based conformational response differences upon binding.

Finally, CPES fuses the static and curvature-informed dynamic representations at the graph level, i.e.,
\begin{equation}
\label{eq:four_g}
\mathbf{g}
=
\mathbf{g}_{\text{static}}^{(\text{lig})}
+
\mathbf{g}_{\text{static}}^{(\text{pro})}
+
\mathbf{g}_{\text{CurvPES}}^{(\text{lig})}
+
\mathbf{g}_{\text{CurvPES}}^{(\text{pro})}.
\end{equation}
A final MLP-based regressor maps the fused representation to the predicted binding affinity $\hat{y}=f(\mathbf{g})$. The model is trained by minimizing the mean squared error loss ${{\mathcal{L}}_{\text{mse}}}=\frac{1}{N}\sum\nolimits_{n=1}^{N}{{{({{{\hat{y}}}_{n}}-{{y}_{n}})}^{2}}}$, with $y_n$ and $\hat y_n$ denoting the experimental and predicted binding affinities of the $n$-th sample, respectively.

\subsection{Implementation Details}
We implement CPES based on the PyTorch Geometric (PyG) framework, with PyTorch serving as the backend. Experiments are conducted on NVIDIA GeForce RTX 4080 GPU with 16 GB of memory. The model parameters are optimized using the Adam optimizer with an initial learning rate of $1 \times 10^{-4}$ and a weight decay of $1 \times 10^{-6}$ to mitigate overfitting. To further stabilize the training process, we employ a learning rate scheduler, which monitors the validation performance and reduces the learning rate by a factor of $0.5$ if no improvement is observed for $50$ consecutive epochs. Source codes are available at https://github.com/Peng-Fei-Sun/CPES.

\section{Experiments}

In this section, we first assess cross-dataset generalization performance on widely used PDBbind benchmarks and the external CSAR NRC-HiQ dataset. We then conduct ablation studies to examine the contributions of the curvature-informed PES descriptors and attention modules, followed by interpretability analyses that reveal how CPES captures both key binding sites and binding-related dynamic modes. Due to the limited space, additional experimental settings and extended results are provided in  Supplement~\ref{appendix:experiment}.

\begin{figure*}[t]
\centerline{\includegraphics[width=0.8\linewidth]{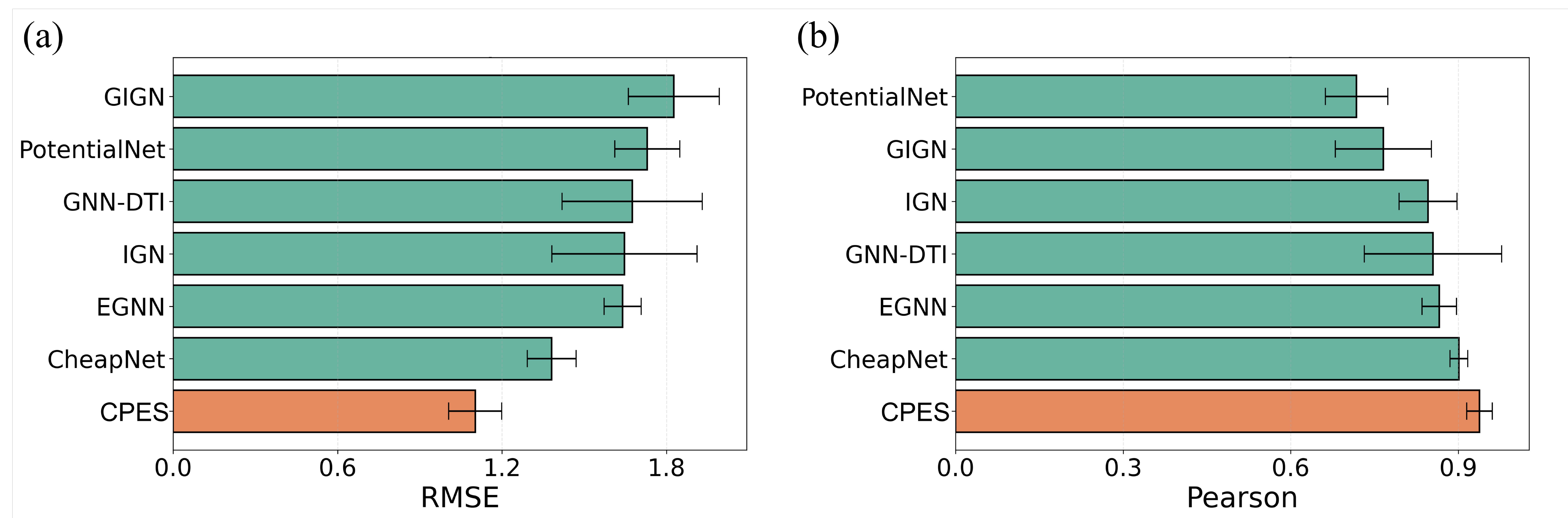}}
\caption{Performance comparison of CPES on the CSAR NRC-HiQ dataset with (a) RMSE and (b) Pearson.}
\label{fig:csar_RMSE_Pearson}
\end{figure*}

\begin{table*}[t]
\centering
\caption{Ablation Study Results of CPES with Varying Settings on the 2013 core set, 2016 core set, and the 2019 holdout set}
\resizebox{\linewidth}{!}{%
\begin{tabular}{lcccccccccc}
\toprule
\multirow{2}{*}{Model} & \multirow{2}{*}{PES} & \multirow{2}{*}{Attn-S} & \multirow{2}{*}{Attn-D} 
& \multicolumn{2}{c}{2013 core} 
& \multicolumn{2}{c}{2016 core} 
& \multicolumn{2}{c}{2019 holdout} \\
\cmidrule(lr){5-6} \cmidrule(lr){7-8} \cmidrule(lr){9-10}
& & & & RMSE $\downarrow$ & Pearson $\uparrow$ 
& RMSE $\downarrow$ & Pearson $\uparrow$ 
& RMSE $\downarrow$ & Pearson $\uparrow$ \\
\midrule
Variant 1 & \ding{55} & \ding{55} & \ding{55} 
& 1.322 (0.051) & 0.845 (0.011) 
& 1.153 (0.045) & 0.855 (0.009) 
& 1.359 (0.026) & 0.663 (0.011) \\
Variant 2 & \ding{55} & \ding{51} & \ding{55} 
& 1.261 (0.023) & 0.859 (0.008) 
& 1.124 (0.018) & 0.861 (0.005) 
& 1.350 (0.013) & 0.664 (0.007) \\
Variant 3 & \ding{51} & \ding{55} & \ding{55} 
& 1.256 (0.019) & 0.859 (0.006) 
& 1.120 (0.028) & 0.863 (0.011) 
& 1.356 (0.025) & 0.666 (0.013) \\
Variant 4 & \ding{51} & \ding{51} & \ding{55} 
& 1.233 (0.032) & 0.868 (0.013) 
& 1.129 (0.009) & 0.859 (0.008) 
& 1.344 (0.011) & 0.670 (0.014) \\
Variant 5 & \ding{51} & \ding{55} & \ding{51} 
& 1.241 (0.021) & 0.861 (0.020) 
& 1.134 (0.023) & 0.868 (0.015) 
& 1.347 (0.009) & 0.662 (0.011) \\
CPES      & \ding{51} & \ding{51} & \ding{51} 
& \textbf{1.193 (0.010)} & \textbf{0.876 (0.006)} 
& \textbf{1.101 (0.009)} & \textbf{0.870 (0.003)} 
& \textbf{1.321 (0.011)} & \textbf{0.678 (0.007)} \\
\bottomrule
\end{tabular}
}
\flushleft
\footnotesize
PES: curvature-informed descriptors derived from the potential energy surface (PES); \\
Attn-S: attention for the static (S) structural interactions; \\
Attn-D: attention for the dynamic (D) conformational modes.
\label{table:ablation_study}
\end{table*}

\subsection{Cross-Dataset Generalization}

\textbf{Datasets.} For a fair comparison, we strictly follow the experimental protocols used in prior works. We adopt the PDBbind dataset and use the same data splits as in GIGN \cite{NEID1} and CheapNet \cite{NEID2}. All baseline models as well as our proposed CPES are trained and evaluated under this unified data splitting scheme, ensuring a strictly controlled and fair comparison across different methods. This setup enables a reliable assessment of cross-dataset generalization performance.

We train the model on 11,903 samples from the PDBbind 2016 general set, with 1,000 samples used for validation. The evaluation is conducted on three widely used benchmark test sets: (i) the CASF-2013 benchmark (107 samples), (ii) the CASF-2016 benchmark (285 samples), both derived from the PDBbind core set, and (iii) a PDBbind 2019 holdout test set consisting of 4,366 complexes that are non-overlapping with the aforementioned splits.

\textbf{Comparison with Baselines.} We compare CPES with a diverse set of representative baselines, covering interaction-free methods, interaction-based methods, and interaction-based methods with attention mechanisms. The interaction-free category includes DeepDTA \cite{NEID18}, GraphDTA (with GCN, GAT, GIN, and GAT-GCN) \cite{NEID19}, MGraphDTA \cite{NEID20}. The interaction-based category comprises RF-Score \cite{NEID106}, Pafnucy \cite{NEID15}, OnionNet \cite{NEID108}, PotentialNet \cite{NEID21}, SchNet \cite{NEID22}, GNN-DTI \cite{NEID23}, IGN \cite{NEID24}, EGNN \cite{NEID25}, GIGN \cite{NEID1}, MetalProGNet \cite{NEID109}, SS-GNN \cite{NEID86}, and EHIGN \cite{NEID91}. The interaction-based methods with attention mechanisms include AttentionSiteDTI \cite{NEID26}, CAPLA \cite{NEID27}, GAABind \cite{NEID28}, DEAttentionDTA \cite{NEID29}, and CheapNet \cite{NEID2}. These methods further introduce attention-based designs to enhance the modeling of important interaction patterns or informative regions. Our proposed CPES belongs to this family but further integrates curvature-informed dynamic descriptors, enabling more effective modeling of both static interaction patterns and binding-related conformational compatibility. The results are summarized in Table~\ref{table:Cross-Dataset}. As shown, CPES consistently achieves the best overall performance across the three benchmark test sets.

\subsection{External Evaluation on Non-PDBbind Test Set}

We further evaluate CPES on the CSAR NRC-HiQ dataset \cite{NEID110}, which serves as an external benchmark for protein-ligand binding affinity prediction outside the PDBbind dataset. Following the evaluation protocol of CheapNet \cite{NEID2}, we remove complexes that cannot be processed by RDKit and exclude those overlapping with the training data, resulting in 14 samples for evaluation. Fig.~\ref{fig:csar_RMSE_Pearson} presents the comparison between CPES and other interaction-based methods.

CPES achieves the best performance among all compared models, with an RMSE of 1.102 and a Pearson correlation coefficient of 0.938, consistently outperforming existing approaches on both metrics. These results highlight the strong generalization capability of CPES, which can be attributed to its curvature-informed inductive bias on conformational dynamics, enabling effective modeling of complex protein-ligand interactions in external datasets.

\begin{figure*}[t]
    \centering
    \includegraphics[width=1\linewidth]{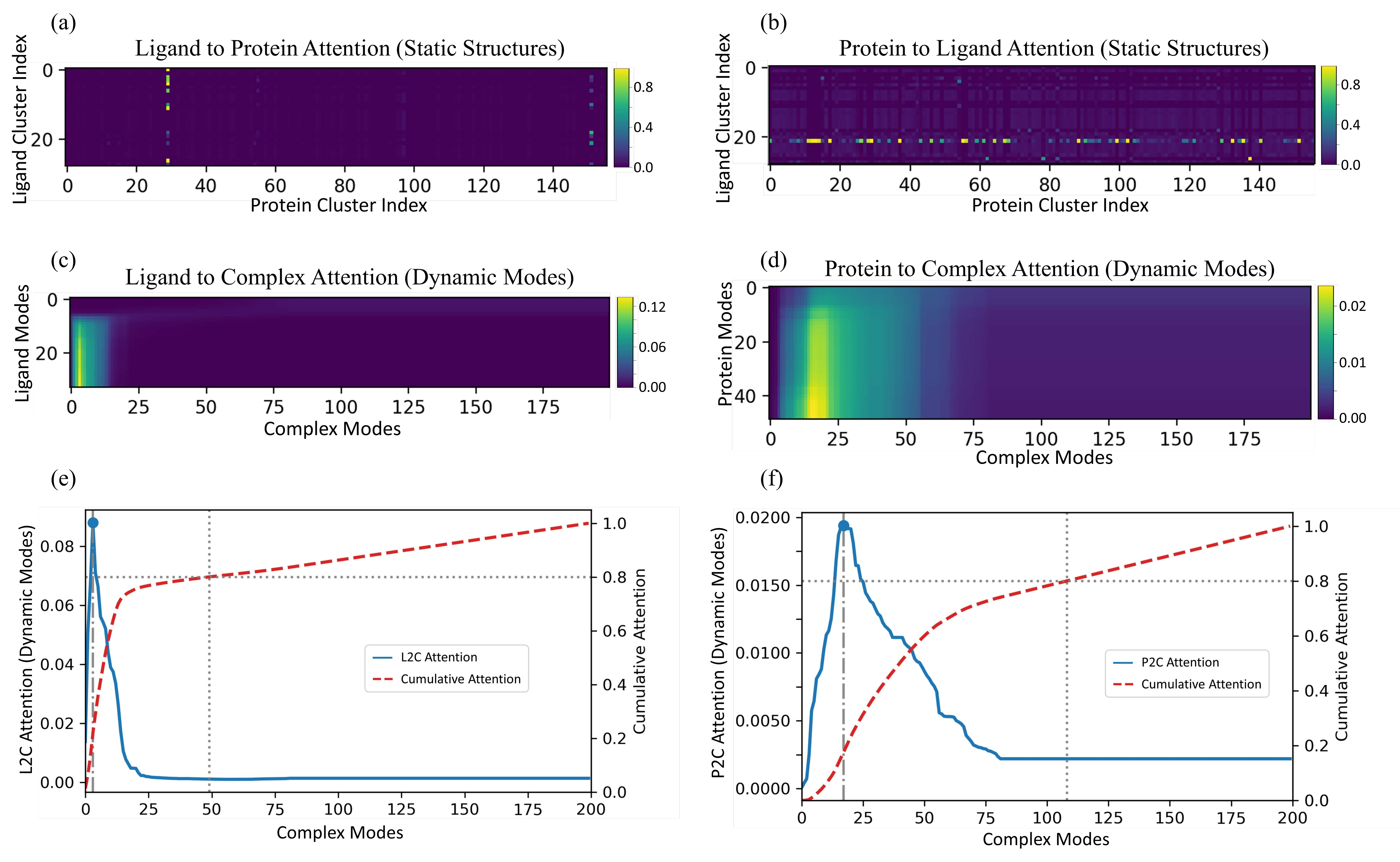}
    \caption{Visualization of cross-attention across static structures and dynamic modes for the complex with PDB ID 3myg. (a,b) Bidirectional cross-attention maps between ligand and protein clusters in the static structural representation. (c,d) Cross-attention maps from ligand modes and protein modes to complex modes in the dynamic representation. (e,f) Mean attention distributions across complex modes for ligand-to-complex (L2C) and protein-to-complex (P2C) attention, together with their cumulative attention curves. In both cases, the attention is mainly concentrated on the low-index complex modes, suggesting that the model preferentially focuses on a small number of dominant collective motions.}
    \label{fig:interpretability_attention_maps}
\end{figure*}

\subsection{Ablation Study}

To evaluate the contribution of each component in CPES, we conduct an ablation study by systematically removing the curvature-informed descriptors derived from the potential energy surface (PES), the static attention module (Attn-S) that captures structure-based interactions at the protein–ligand interface, and the dynamic attention module (Attn-D) that models conformational flexibility through spectral representations of collective modes. The results on the 2013 core set, 2016 core set, and the 2019 holdout set are summarized in Table~\ref{table:ablation_study}.

Overall, CPES consistently achieves the best performance across all datasets, demonstrating the effectiveness of jointly modeling static interactions and dynamic information under the conformational curvature-informed inductive bias. First, removing all three components (Variant\,1) leads to the worst performance, indicating that the baseline model without these mechanisms is insufficient to capture the complexity of protein-ligand binding. Introducing the static attention module alone (Variant\,2) yields noticeable improvements over Variant\,1, suggesting that modeling static interactions at the binding interface is beneficial. Similarly, incorporating only the PES descriptors (Variant\,3) also improves performance, highlighting the importance of curvature-informed geometric features. When both PES and Attn-S are included (Variant\,4), the performance is further enhanced, indicating that curvature-aware representations and static interaction modeling are complementary. On the other hand, introducing the dynamic attention module (Variant\,5) leads to additional gains, demonstrating the value of capturing conformational flexibility through spectral representations. Notably, the incorporation of PES consistently improves performance across all variants, indicating that curvature-informed descriptors provide meaningful inductive bias beyond purely geometric representations. Moreover, the dynamic attention module further enhances performance, particularly on the 2019 holdout set, suggesting its effectiveness in capturing binding-induced flexibility and improving generalization to unseen complexes.

\begin{figure*}[t]
    \centering
    \includegraphics[width=1\linewidth]{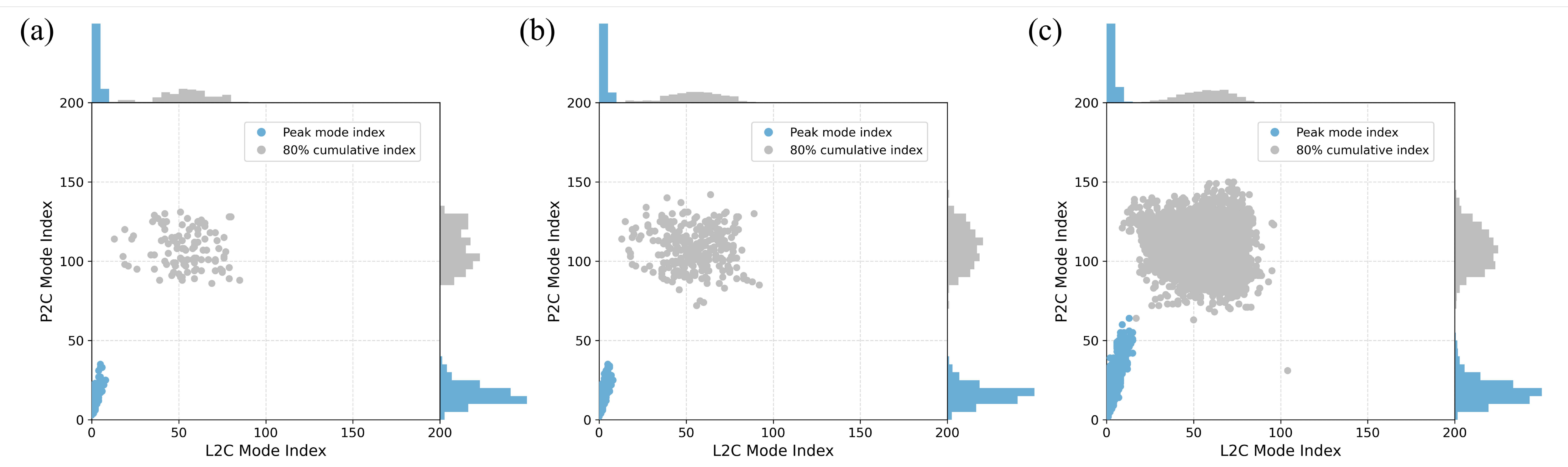}
    \caption{Scatter plots of the attention mode indices for the L2C and P2C dynamic modes cross-attention on the (a) 2013 core set, (b) 2016 core set, and (c) 2019 holdout set. Blue points denote the peak mode indices of the mean attention distributions, while gray points denote the mode indices at which the cumulative attention reaches 80\%. Marginal histograms on the top and right show the corresponding distributions along the L2C and P2C axes, respectively.}
    \label{fig:attentionpeakscatter}
\end{figure*}

\subsection{Interpretability of CPES}

To investigate whether the improved performance of CPES is accompanied by meaningful physical interpretability, we analyze the cross-attention patterns in both the static structural space and the dynamic modal space. We explicitly combines structural interaction learning with conformational curvature-informed dynamical descriptors. The attention maps offer a direct way to examine how these two information streams contribute to affinity prediction.

Fig.~\ref{fig:interpretability_attention_maps}(a) and (b) illustrate the bidirectional cross-attention between ligand and protein clusters in the static structural representation. The horizontal and vertical axes correspond to cluster indices, since atoms are grouped into clusters via soft clustering in the preceding representation learning stage. Fig.~\ref{fig:interpretability_attention_maps}(a) shows the ligand-to-protein attention, where each row of the heatmap is normalized to sum to one, indicating how each ligand cluster distributes its attention over protein clusters. Fig.~\ref{fig:interpretability_attention_maps}(b), on the other hand, presents the protein-to-ligand attention, where each column is normalized, reflecting how each protein cluster allocates attention to ligand clusters. The cross-attention maps of static structural representation is sparse and nonuniform, with only a limited number of cluster pairs receiving strong responses. This indicates that the static structural component of CPES identifies a subset of structurally important interaction regions at the binding interface. Such selective emphasis is desirable, since protein-ligand recognition is typically dominated by a limited number of critical local interactions.

By comparison, Fig.~\ref{fig:interpretability_attention_maps}(c) and (d) exhibit a notably different pattern in the dynamic modal space. For clearer visualization and compact figure presentation, the padded ligand and protein modes are removed. Instead of the discrete, particle-like distributions observed in Fig.~\ref{fig:interpretability_attention_maps}(a, b), the attention maps in Fig.~\ref{fig:interpretability_attention_maps}(c, d) appear smoother and more gradually varying along the complex mode axis, with attention predominantly concentrated in the low-index modal region and then progressively decaying toward higher modes. In other words, while the static attention mainly identifies localized structural binding sites, the dynamic attention reflects a broader distribution over the modal spectrum and emphasizes a small set of dominant low-frequency collective motions. This difference indicates that the two branches of CPES provide complementary interpretability: the static branch identifies key structural interaction regions, whereas the dynamic branch characterizes binding-relevant dynamical coordination at the level of collective modes.

Fig.~\ref{fig:interpretability_attention_maps}(e) and (f) aggregate the attention contributions from ligand modes or protein modes to each complex mode, resulting in mean attention distributions over the complex mode spectrum. This mean attention can be interpreted as the overall level of attention that the unbound ligand or protein collectively assigns to each mode of the bound complex. This compression reduces the detailed pairwise information in the original attention maps, but offers a clearer view of the overall distribution trend. In addition, the cumulative attention curves further illustrate how the attention mass accumulates across the complex modes. For both ligand-to-complex and protein-to-complex attention, the mean attention distributions exhibit peaks in the low-index region of the complex mode spectrum, while the cumulative attention curves increase rapidly at early modes and then flatten progressively. These trends show that most of the dynamical attention scores are captured by only a small number of low-index complex modes. Since low-index modes usually correspond to slow, large-scale, and cooperative motions, this suggests that CPES preferentially exploits the dominant collective motions that are most likely to be relevant to conformational adaptation and molecular recognition.

To further summarize this behavior across test samples, we plot the peak attention mode index and the 80\% cumulative attention index of the ligand-to-complex and protein-to-complex attentions in Fig.~\ref{fig:attentionpeakscatter}. The blue points represent the peak modes receiving the maximum mean attention, and the gray points represent the mode indices at which the cumulative attention reaches 80\%. It can be observed that the peak indices are consistently concentrated near the lowest-index region, indicating that the strongest dynamic responses are dominated by early complex modes. In addition, the scatter distributions exhibit a bias toward smaller L2C indices, suggesting stronger preference of the ligand-to-complex attention for low-index complex modes. Meanwhile, the 80\% cumulative indices remain distributed within a relatively limited spectral range, suggesting that most of the attention mass can already be captured by only a subset of low-frequency complex modes rather than being broadly distributed across the entire spectrum. The marginal distributions further support this observation, showing that both the peak indices and the 80\% cumulative indices remain concentrated within the low-frequency region of the complex mode spectrum. These results provide sample-level evidence that CPES systematically emphasizes dominant collective motions of the bound complex during cross-attention for dynamic modes. 

\section{Conclusion}

In this work, we proposed CPES, a curvature-informed potential energy surface graph neural network for protein-ligand binding affinity prediction. By introducing Hessian eigenspectrum descriptors derived from the anisotropic network model (ANM), our model incorporates curvature-informed PES representations into molecular representation learning. Extensive experiments across diverse benchmark datasets demonstrate that CPES achieves competitive predictive performance together with robust generalization capability. Ablation studies further confirm the effectiveness of the PES curvature descriptors and both the static and dynamic attention modules. Moreover, interpretability analyses reveal that the static branch identifies sparse binding-relevant interaction regions, whereas the dynamic branch consistently focuses on low-frequency collective modes of the bound complex, highlighting the importance of collective conformational dynamics in protein-ligand binding. The results demonstrate that PES curvature provides an effective physical inductive bias by characterizing molecular conformational flexibility and collective dynamics beyond static interaction geometry. In future work, we will explore more accurate physical energy models and broader applications of curvature-informed potential energy surface representations in structure-based drug discovery tasks, such as flexible docking and virtual screening.



\section*{References}

\putbib[myrefs]
\end{bibunit}

\clearpage
\twocolumn[
\begin{center}
\LARGE \textbf{Curvature-Informed Potential Energy Surface for Protein-Ligand Binding Affinity Prediction\\ (Supplementary Information)}
\end{center}
\vspace{3.5em}]

\setcounter{section}{0}
\renewcommand{\theHsection}{supp.section.\arabic{section}}
\renewcommand{\thetable}{S\arabic{table}}
\renewcommand{\theHtable}{S\arabic{table}}
\setcounter{table}{0}
\renewcommand{\thefigure}{S\arabic{figure}}
\renewcommand{\theHfigure}{S\arabic{figure}}
\setcounter{figure}{0}
\begin{bibunit}
\nocite{IEEEexample:BSTcontrol}

This supplementary information provides additional discussions, theoretical derivations, architectural details, and extended experimental results that complement the main manuscript. A broader discussion of related works is included, covering molecular dynamics and energy-based approaches, geometric deep learning methods for protein-ligand binding affinity prediction, as well as classical conformational dynamics methods such as normal mode analysis (NMA) and anisotropic network models (ANM), together with additional references. Detailed derivations of the SE(3) equivariance property of the Hessian and the invariance of its eigenvalue spectrum under rigid-body transformations are presented. Additional descriptions of the CPES framework are also provided, particularly for the geometry-aware static interaction branch. More experimental details and extended results are further included, including dataset processing procedures, evaluation settings, experiments on dissimilar protein scenarios, and analyses of different graph construction strategies.

\section{Additional Discussion on Related Works}
\label{appendix:Related Works}

\subsection{Molecular Dynamics and Energy-Based Approaches}

Molecular dynamics (MD) simulations constitute a fundamental class of physics-based approaches for modeling protein-ligand binding by explicitly describing atomic motions under physically grounded force fields. By integrating Newtonian equations of motion, MD enables the exploration of conformational ensembles and provides mechanistic insights into binding phenomena such as induced fit, conformational selection, and binding-unbinding pathways \cite{NEID30,NEID31}. Beyond classical MD, several complementary energy-based approaches have been developed for protein-ligand docking and binding affinity prediction \cite{NEID33}. Quantum chemical methods, including density functional theory (DFT), model protein-ligand interactions by explicitly describing electronic structure and polarization effects, but their high computational cost limits their applicability to small systems or localized regions \cite{NEID34,NEID35}. Despite their physical rigor, these energy-based methods typically require substantial computational resources and long simulation timescales, limiting their practicality for large-scale screening or dataset-level affinity prediction \cite{NEID36,NEID37,NEID38,NEID39}. This limitation motivates the development of deep learning-based approaches that incorporate physically meaningful energy-inspired cues while maintaining scalability and efficiency.

\subsection{Geometric Deep Learning for Protein-Ligand Binding Affinity Prediction}

Geometric deep learning (GDL) has emerged as a powerful paradigm for learning from structured data by generalizing neural networks to non-Euclidean domains such as graphs, manifolds, and meshes \cite{NEID40}. Within this framework, graph neural networks (GNNs) have become the de facto choice for learning with graph-structured data \cite{NEID45,NEID46,NEID47,NEID48,NEID49}, owing to their ability to model relational information through message passing while incorporating geometric priors and symmetry properties. By explicitly encoding invariance or equivariance to transformations such as translation, rotation, and permutation \cite{NEID50,NEID51,NEID52,NEID53,NEID54}, geometric deep learning architectures achieve improved accuracy and data efficiency, particularly in physically grounded domains.

Motivated by these properties, geometric deep learning has been widely adopted for protein-ligand binding affinity prediction and other molecular modelling \cite{NEID41,NEID42,NEID43,NEID44}. Graph-based models represent protein-ligand complexes as three-dimensional graphs, where nodes correspond to atoms and edges encode covalent or non-covalent interactions, enabling explicit modeling of spatial relationships and atomic interactions. Compared with sequence-based or descriptor-driven approaches, such interaction-based models introduce inductive biases that are more closely aligned with physical binding mechanisms and have demonstrated superior performance across diverse benchmarks. Representative frameworks include interaction graph neural networks (IGNNs), such as GIGN \cite{NEID1}, which integrate physically meaningful interactions into invariant message-passing schemes, as well as hierarchical models like CheapNet \cite{NEID2} that aggregate atom-level representations into cluster-level interactions via differentiable pooling and cross-attention. Despite these advances, most existing geometric deep learning approaches still rely on static representations derived from a single bound conformation, implicitly treating protein-ligand complexes as rigid entities and overlooking conformational flexibility and binding-induced dynamics. This limitation motivates the development of models that incorporate physically informed inductive biases beyond static interaction geometry.

\subsection{Inductive Bias from Conformational Energy Distributions}

Inductive bias plays a central role in machine learning by shaping how models generalize beyond observed data \cite{NEID56,NEID57,NEID59}. For protein-ligand binding affinity prediction, inductive biases that reflect underlying physical and biological mechanisms are particularly important, as they encourage representations that capture causal structure-function relationships rather than superficial correlations, among which conformational energy distributions play a central role in shaping molecular stability and adaptability \cite{NEID60,NEID61,NEID62}.

From a biophysical perspective, protein-ligand binding does not occur between rigid structures but rather involves ensembles of conformations characterized by an underlying potential energy function. The accessibility of binding-competent states and binding-induced conformational changes is governed by the local curvature of the energy function in the vicinity of stable conformations. While explicitly sampling full conformational energy distributions using molecular dynamics is computationally prohibitive at scale, the local behavior of the potential energy function already encodes essential information about molecular flexibility. Classical approaches such as normal mode analysis (NMA) \cite{NEID63,NEID64,NEID65,NEID66,NEID67,NEID68} and elastic network models \cite{NEID69,NEID70,NEID71,NEID72} leverage this observation by approximating the local curvature of the potential energy function around equilibrium conformations. By analyzing second-order variations of the energy function, these methods characterize intrinsic stiffness, collective motions, and dominant deformation modes that are closely related to binding-relevant conformational changes. Importantly, such local energy-derived descriptors provide compact, physically grounded summaries of molecular flexibility without requiring extensive sampling.

Motivated by these principles, incorporating representations inspired by conformational energy distributions introduces an inductive bias that aligns learning objectives with physical binding mechanisms \cite{NEID55,NEID73,NEID74}. Rather than relying solely on static interaction geometry, energy-informed inductive biases enable models to capture how molecular systems respond to perturbations and adapt during binding, that is, conformational dynamic differences between bound and unbound states, thereby improving both generalization and interpretability in protein-ligand binding affinity prediction.

\section{Proof of SE(3) Equivariance of Hessian}
\label{appendix:proof}
Let $\mathbf q\in\mathbb R^{3N}$ be the configuration vector and $V:\mathbb R^{3N}\to\mathbb R$ the potential energy function. Consider an arbitrary rotation matrix $\mathbf{R} \in \mathrm{SO}(3)$ and a translation vector $\mathbf t\in\mathbb R^3$. Define the block-diagonal matrix
\begin{equation*}
\mathbf{Q} = \operatorname{diag}(\mathbf{R}, \ldots, \mathbf{R}) \in \mathbb{R}^{3N \times 3N},
\end{equation*}
and the translation vector
\begin{equation*}
\mathbf{c} = (\mathbf{t}, \ldots, \mathbf{t}) \in \mathbb{R}^{3N}.
\end{equation*}
The global rigid-body transformation is given by
\begin{equation*}
\mathbf{q}' = \mathbf{Q}\mathbf{q} + \mathbf{c}.
\end{equation*}

Assume that the potential energy depends only on the internal geometric relations of the system and is therefore invariant under such a transformation, i.e.,
\begin{equation*}
V'(\mathbf{q}') \equiv V(\mathbf{q}).
\end{equation*}
Since $\mathbf{R} \in \mathrm{SO}(3)$, the induced matrix $\mathbf{Q}$ is orthogonal and satisfies $\mathbf{Q}^{\top} \mathbf{Q} = \mathbf{I}$. Hence the transformation is invertible, with inverse
\begin{equation*}
\mathbf{q} = \mathbf{Q}^{\top} (\mathbf{q}' - \mathbf{c}).
\end{equation*}
Consequently, $V(\mathbf q)$ can be regarded as the composite function $V(\mathbf q(\mathbf q'))$.

Taking the gradient of both sides of the identity $V'(\mathbf q')\equiv V(\mathbf q(\mathbf q'))$ with respect to $\mathbf q'$, the left-hand side yields $\nabla_{\mathbf q'}V'(\mathbf q')$, while the right-hand side is a scalar-valued composite function. By the chain rule for scalar functions (with the convention that gradients are column vectors), it follows that
\begin{equation*}
\nabla_{\mathbf{q}'} V\bigl(\mathbf{q}(\mathbf{q}')\bigr)=
\left( \frac{\partial \mathbf{q}}{\partial \mathbf{q}'} \right)^{\!\top}
\nabla_{\mathbf{q}} V(\mathbf{q}).
\end{equation*}
From $\mathbf{q} = \mathbf{Q}^{\top} (\mathbf{q}' - \mathbf{c})$, the Jacobian matrix is constant and given by
\begin{equation*}
\frac{\partial \mathbf{q}}{\partial \mathbf{q}'} = \mathbf{Q}^{\top},
\end{equation*}
where the translation vector $\mathbf c$ vanishes upon differentiation. Substituting this result yields
\begin{equation*}
\nabla_{\mathbf{q}'} V'(\mathbf{q}') = \mathbf{Q}\,\nabla_{\mathbf{q}} V(\mathbf{q}).
\end{equation*}

Differentiating once more with respect to $\mathbf q'$ gives
\begin{equation*}
\nabla_{\mathbf q'}^{2}V'(\mathbf q')=
\nabla_{\mathbf q'}\!\left(\mathbf Q\,\nabla_{\mathbf q}V(\mathbf q)\right).
\end{equation*}
Since $\mathbf Q$ is independent of $\mathbf q'$, it can be taken outside the derivative 
\begin{equation*}
\nabla_{\mathbf q'}^{2}V'(\mathbf q')=
\mathbf Q\,\nabla_{\mathbf q'}\!\left(\nabla_{\mathbf q}V(\mathbf q)\right).
\end{equation*}
Note that $\nabla_{\mathbf q}V(\mathbf q)$ is a vector-valued function from $\mathbb R^{3N}$ to $\mathbb R^{3N}$, and its first derivative with respect to $\mathbf q'$ is a Jacobian matrix. Applying the chain rule for vector-valued composite functions, the following is derived 
\begin{equation*}
\nabla_{\mathbf q'}\!\left(\nabla_{\mathbf q}V(\mathbf q(\mathbf q'))\right)=
\left(\nabla_{\mathbf q}^{2}V(\mathbf q)\right)
\frac{\partial \mathbf q}{\partial \mathbf q'}.
\end{equation*}
Substituting $\frac{\partial \mathbf q}{\partial \mathbf q'} = \mathbf Q^\top$ yields
\begin{equation*}
\nabla_{\mathbf q'}^{2}V'(\mathbf q')=
\mathbf Q\left(\nabla_{\mathbf q}^{2}V(\mathbf q)\right)\mathbf Q^{\top}.
\end{equation*}

Defining $\mathbf H(\mathbf q)=\nabla_{\mathbf q}^{2}V(\mathbf q)$ and $\mathbf H'(\mathbf q')=\nabla_{\mathbf q'}^{2}V'(\mathbf q')$, we finally obtain 
\begin{equation*}
    \mathbf H'(\mathbf q')=\mathbf Q\,\mathbf H(\mathbf q)\,\mathbf Q^{\top}.
\end{equation*}
This completes the proof of SE(3) equivariance of the Hessian.

\section{Details of the CPES Framework}
\label{appendix:framework_static}

Given a complex graph $\mathcal{G}=(\mathcal{V},\mathcal{E})$ the node set is defined as $\mathcal V=\mathcal V_{\text{lig}}\cup\mathcal V_{\text{pro}}$, and the edge set as $\mathcal E=\mathcal E_{\mathrm{intra}}\cup\mathcal E_{\mathrm{inter}}$ (where ${{\mathcal{E}}_{\mathrm{intra}}}={{\mathcal{E}}_{\text{lig}}}\cup {{\mathcal{E}}_{\text{pro}}}$), representing intra-molecular and inter-molecular interactions, respectively. Each node $i\in\mathcal V$ is associated with an initial feature vector $\mathbf x_i$ and a three-dimensional coordinate $\mathbf r_i$. Node features are first embedded into a $d$-dimensional hidden space via a learnable embedding function
\begin{equation*}
\mathbf{h}_i^{(0)} = \operatorname{MLP}_{\text{emb}}(\mathbf{x}_i) \in \mathbb{R}^d.
\end{equation*}

In the static structural branch, geometry-aware message passing is performed over both edge types $\mathcal E_r$ with $r\in\{\mathrm{intra},\mathrm{inter}\}$. At layer $t$, a channel-wise modulation vector induced by relative geometric relations is defined as
\begin{equation*}
\boldsymbol{\rho}_{ij}^{(r)} = \operatorname{\psi}^{(r)}\!\left( \lVert \mathbf{r}_i - \mathbf{r}_j \rVert \right) \in \mathbb{R}^d,
\end{equation*}
where $\psi^{(r)}(\cdot)$ is a learnable mapping that transforms inter-node distances into channel-wise modulation coefficients. The aggregated message for node $i$ under edge type $r$ is given by
\begin{equation*}
\mathbf{m}_i^{(r,t)} = \sum_{j : (j,i) \in \mathcal{E}_r} \mathbf{h}_j^{(t)} \odot \boldsymbol{\rho}_{ji}^{(r)},
\end{equation*}
where $\odot$ denotes the element-wise product, enabling geometry-aware modulation of neighbor features. Node representations are then updated via a MLP transformation with residual connection, i.e., 
\begin{equation*}
\begin{aligned}
\mathbf{h}_i^{(\mathrm{intra},t+1)}
&=
\operatorname{MLP}_{\mathrm{intra}}\!\left(
\mathbf{m}_i^{(\mathrm{intra},t)}
\right)
+
\mathbf{h}_i^{(t)}, \\
\mathbf{h}_i^{(\mathrm{inter},t+1)}
&=
\operatorname{MLP}_{\mathrm{inter}}\!\left(
\mathbf{m}_i^{(\mathrm{inter},t)}
\right)
+
\mathbf{h}_i^{(t)}.
\end{aligned}
\end{equation*}
The updates from intra-molecular and inter-molecular edges are fused at the node level as
\begin{equation*}
\mathbf{h}_i^{(t+1)} = \tfrac{1}{2}\left( \mathbf{h}_i^{(\mathrm{intra},\,t+1)} + \mathbf{h}_i^{(\mathrm{inter},\,t+1)} \right).
\end{equation*}

To extract higher-level, hierarchically organized representations from atom-level features, differentiable clustering pooling is applied separately to the ligand and protein subgraphs. For $s\in\{\text{lig},\text{pro}\}$, let
\begin{equation*}
\mathbf F_s\in\mathbb R^{N_s\times d},\qquad 
\mathbf A_s\in\mathbb R^{N_s\times N_s},
\end{equation*}
denote the node feature matrix and adjacency matrix, respectively. $N_s$ denotes the number of nodes in the ligand or protein subgraph. A soft assignment matrix is produced by a graph neural network (GNN) 
\begin{equation*}
\mathbf{S}_s = \operatorname{softmax}\!\left( \operatorname{GNN}_s(\mathbf{F}_s, \mathbf{A}_s) \right) \in \mathbb{R}^{N_s \times C_s},
\end{equation*}
where $C_s$ denotes the number of clusters for subgraph $s$. Cluster-level representations and adjacencies are then constructed as
\begin{equation*}
\begin{aligned}
\mathbf{Z}_s &= \mathbf{S}_s^\top \mathbf{F}_s \in \mathbb{R}^{C_s \times d}, \\
\widetilde{\mathbf{A}}_s &= \mathbf{S}_s^\top \mathbf{A}_s \mathbf{S}_s \in \mathbb{R}^{C_s \times C_s}.
\end{aligned}
\end{equation*}
After further updates on the cluster graphs, cluster-level representations $\mathbf Z_{\text{lig}}$ and $\mathbf Z_{\text{pro}}$ are obtained.

At the cluster level, bidirectional cross-attention is employed to model static structural interactions between the ligand and the protein. The ligand-to-protein interaction is computed as
\begin{equation*}
\begin{aligned}
\widetilde{\mathbf{Z}}_{\text{lig} \to \text{pro}}
&= \operatorname{softmax}\!\left(
\frac{\mathbf{Z}_{\text{lig}} \mathbf{W}^{(\text{l2p})}
\left( \mathbf{Z}_{\text{pro}} \mathbf{W}^{(\text{l2p})} \right)^\top}{\sqrt{d}}
\right) \\
&\quad \cdot \mathbf{Z}_{\text{pro}} \mathbf{W}^{(\text{l2p})}
\in \mathbb{R}^{C_{\text{lig}} \times d},
\end{aligned}
\end{equation*}
which is aggregated into a graph-level representation with a residual connection 
\begin{equation*}
\mathbf{g}_{\text{static}}^{(\text{lig})}
=
\operatorname{MLP}_{\text{out}}^{(\text{l2p})}\!\left(
\sum_{i=1}^{C_{\text{lig}}} \widetilde{\mathbf{Z}}_{\text{lig} \to \text{pro}, i}
\right)
+
\sum_{i=1}^{C_{\text{lig}}} \mathbf{Z}_{\text{lig}, i}.
\end{equation*}
Symmetrically, the protein-to-ligand interaction is given by
\begin{equation*}
\begin{aligned}
\widetilde{\mathbf{Z}}_{\text{pro} \to \text{lig}}
&= \operatorname{softmax}\!\left(
\frac{\mathbf{Z}_{\text{pro}} \mathbf{W}^{(\text{p2l})}
\left( \mathbf{Z}_{\text{lig}} \mathbf{W}^{(\text{p2l})} \right)^\top}{\sqrt{d}}
\right) \\
&\quad \cdot \mathbf{Z}_{\text{lig}} \mathbf{W}^{(\text{p2l})}
\in \mathbb{R}^{C_{\text{pro}} \times d},
\end{aligned}
\end{equation*}
and aggregated as
\begin{equation*}
\mathbf{g}_{\text{static}}^{(\text{pro})}
=
\operatorname{MLP}_{\text{out}}^{(\text{p2l})}\!\left(
\sum_{i=1}^{C_{\text{pro}}} \widetilde{\mathbf{Z}}_{\text{pro} \to \text{lig}, i}
\right)
+
\sum_{i=1}^{C_{\text{pro}}} \mathbf{Z}_{\text{pro}, i}.
\end{equation*}
In this way, the bidirectional cross-attention module produces the graph-level static representations $\mathbf{g}_{\text{static}}^{(\text{lig})}$ and $\mathbf{g}_{\text{static}}^{(\text{pro})}$ in Eq.~\eqref{eq:four_g}.

\section{More Experimental Details and Results}
\label{appendix:experiment}

\begin{table*}[t]
\caption{Performance comparison of CPES and baselines on the LBA 30\% and LBA 60\% splits.}
\label{table:LBA}
\centering
\resizebox{\textwidth}{!}{
\begin{tabular}{llcccccc}
\toprule
\multirow{2}{*}{} & \multirow{2}{*}{Model} &
\multicolumn{3}{c}{LBA 30\%} &
\multicolumn{3}{c}{LBA 60\%} \\
\cmidrule(lr){3-5} \cmidrule(lr){6-8}
& & RMSE$\downarrow$ & Pearson$\uparrow$ & Spearman$\uparrow$
& RMSE$\downarrow$ & Pearson$\uparrow$ & Spearman$\uparrow$ \\
\midrule

\multirow{4}{*}{Interaction-free}
& DeepDTA \cite{NEID18} & 1.866 (0.080) & 0.472 (0.022) & 0.471 (0.024) & 1.762 (0.261) & 0.666 (0.012) & 0.663 (0.015) \\
& SSA \cite{NEID111} & 1.985 (0.006) & 0.165 (0.006) & 0.152 (0.024) & 1.891 (0.004) & 0.249 (0.006) & 0.275 (0.008) \\
& TAPE \cite{NEID112} & 1.890 (0.035) & 0.338 (0.044) & 0.286 (0.124) & 1.633 (0.016) & 0.568 (0.033) & 0.571 (0.021) \\
& ProtTrans \cite{NEID113} & 1.544 & 0.438 & 0.434 & 1.641 & 0.595 & 0.588 \\

\midrule

\multirow{5}{*}{Pre-training}
& EGNN-PLM \cite{NEID114} & 1.403 (0.010) & 0.565 (0.020) & 0.544 (0.010) & 1.559 (0.020) & 0.644 (0.020) & 0.646 (0.020) \\
& Uni-Mol \cite{NEID115} & 1.434 & 0.565 & 0.540 & 1.357 & 0.753 & 0.750 \\
& ProFSA \cite{NEID116} & 1.377 (0.010) & 0.628 (0.010) & 0.620 (0.010) & 1.377 (0.010) & 0.764 (0.000) & 0.762 (0.010) \\
& GeoSSL \cite{NEID117} & 1.451 & 0.577 & 0.572 & - & - & - \\
& BindNet \cite{NEID87} & 1.340 & 0.632 & 0.620 & \underline{1.230} & 0.793 & 0.788 \\

\midrule

\multirow{10}{*}{Interaction-based}
& Atom3D-GNN \cite{NEID118} & 1.601 (0.048) & 0.545 (0.027) & 0.533 (0.033) & 1.408 (0.069) & 0.743 (0.022) & 0.743 (0.027) \\
& IEConv \cite{NEID119} & 1.554 (0.016) & 0.414 (0.053) & 0.428 (0.032) & 1.473 (0.024) & 0.667 (0.011) & 0.675 (0.019) \\
& Holoprot \cite{NEID120} & 1.464 & 0.509 & 0.500 & 1.365 & 0.749 & 0.742 \\
& ProNet \cite{NEID121} & 1.463 (0.001) & 0.551 (0.005) & 0.551 (0.008) & 1.343 (0.025) & 0.765 (0.009) & 0.761 (0.003) \\
& LEFTNet \cite{NEID122} & 1.366 (0.016) & 0.592 (0.014) & 0.580 (0.011) & - & - & - \\
& GET \cite{NEID123} & \underline{1.327 (0.005)} & 0.620 (0.004) & 0.611 (0.003) & - & - & - \\
& CheapNet \cite{NEID2} & \textbf{1.311 (0.003)} & \underline{0.642 (0.001)} & \underline{0.639 (0.010)} & 1.238 (0.005) & \underline{0.794 (0.002)} & \underline{0.789 (0.001)} \\
& CPES (ours)& \underline{1.327 (0.017)} & \textbf{0.651 (0.008)} & \textbf{0.641 (0.011)} & \textbf{1.217 (0.004)} & \textbf{0.803 (0.003)} & \textbf{0.802 (0.002)} \\

\bottomrule
\end{tabular}
}
\end{table*}

\begin{figure*}[t]
    \centering
    \includegraphics[width=1\linewidth]{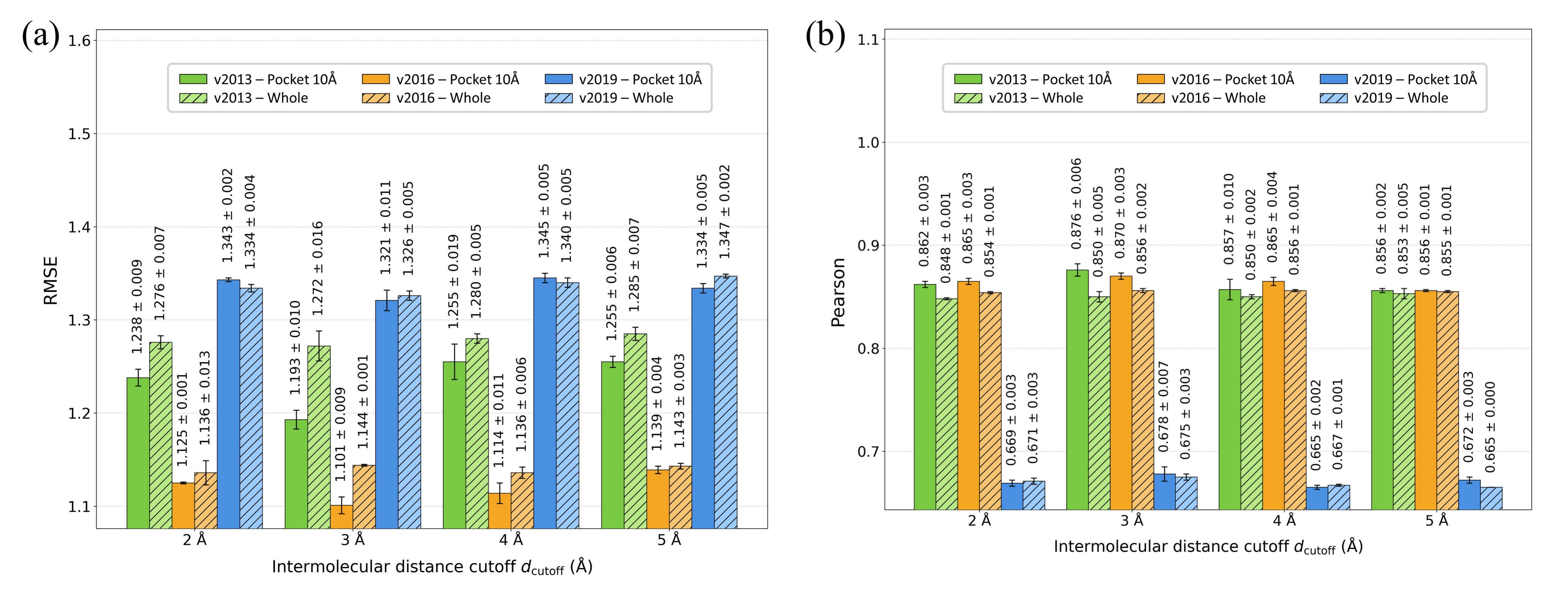}
    \caption{Performance comparison under different intermolecular distance cutoffs $d_{\text{cutoff}}$ and protein representation configurations (10~\AA\ pocket versus whole protein) on the 2013 core set, 2016 core set, and the 2019 holdout set. (a) RMSE and (b) Pearson correlation coefficient are reported for each configuration. Solid bars denote the 10~\AA\ pocket representation, while hatched bars denote the whole protein representation. }
    \label{fig:dcutoff}
\end{figure*}

\subsection{Experimental Setup}
\subsubsection{Datasets Preparation}
Our experiments are mainly conducted on the PDBbind dataset, which contains experimentally resolved 3D structures of protein-ligand complexes along with their binding affinities commonly defined as $-\log\!\left(K_{\text{d}}\right)$ or $-\log\!\left(K_{\text{i}}\right)$, where $K_{\text{d}}$ and $K_{\text{i}}$ correspond to the dissociation and inhibition constants, respectively. The data preprocessing pipeline consists of the following steps. First, we use PyMOL to select protein residues within a distance cutoff $\alpha_{\text{dis}}$ from the ligand. A residue is included if any of its atoms lies within the cutoff, and the entire residue is then retained. Water molecules and hydrogen atoms are removed. Specifically, for constructing the static structural interaction graphs, we set $\alpha_{\text{dis}} = 5\,\text{\AA}$. For the graphs used to derive curvature-informed descriptors of dynamic conformations, we use a larger cutoff $\alpha_{\text{dis}} = 10\,\text{\AA}$. This choice avoids insufficient graph connectivity caused by overly small cutoffs, which may otherwise result in the absence of non-zero eigenmodes in the dynamical spectrum. Then, the ligand molecules are converted from MOL2 to PDB format using Open Babel and subsequently parsed using RDKit. Samples that cannot be processed by RDKit are discarded. Finally, the atomic coordinates, atomic features, and chemical bond information obtained from RDKit are used to construct graph representations. The spectral descriptors of dynamic conformational curvature are then derived from the connectivity structure of the graphs. In our experiments, the numbers of selected modes for the ligand, protein, and complex are set to $K_{\mathrm{ul}} = 200$, $K_{\mathrm{up}} = 200$, and $K_{\mathrm{uc}} = 200$, respectively. These settings retain a sufficient range of low-frequency modes for describing dynamic conformational curvature while maintaining manageable computational complexity in the subsequent cross-attention modeling. For samples with fewer available non-zero modes, the spectral sequence is padded using the last available non-zero eigenvalue to preserve the predefined input length.

\subsubsection{Evaluation Metrics}
To comprehensively evaluate the performance of the proposed model across different datasets, we adopt three widely used regression metrics, including Root Mean Square Error (RMSE), Pearson correlation coefficient, and Spearman rank correlation coefficient. RMSE measures the absolute discrepancy between predicted and ground-truth binding affinities and is particularly sensitive to large errors, thereby reflecting overall prediction accuracy. Pearson evaluates the linear correlation between predictions and true values, capturing the consistency of numerical trends regardless of scale differences. Spearman further assesses the monotonic relationship based on ranked values, providing a more robust measure under potential non-linear dependencies. Together, these metrics offer a comprehensive evaluation from the perspectives of absolute error, linear correlation, and ranking consistency.

\subsection{Dissimilar Protein Scenarios with Atom3D}

\textbf{Datasets.} To further evaluate model performance under dissimilar protein scenarios, experiments are additionally conducted on the Ligand Binding Affinity (LBA) benchmark from the ATOM3D project \cite{NEID118}. The ATOM3D LBA benchmark adopts sequence-identity-based partitioning to reduce similarity between training and test proteins. Following the official benchmark protocol, two evaluation settings are considered: LBA30\% and LBA60\%, corresponding to protein sequence identity thresholds of $30\%$ and $60\%$ respectively. Under these settings, proteins in the test set share at most the specified sequence identity with proteins in the training set, providing a more rigorous evaluation of model robustness. 

\textbf{Comparison with Baselines.} Table~\ref{table:LBA} compares the proposed model with representative interaction-free methods (e.g., TAPE \cite{NEID112} and ProtTrans \cite{NEID113}), pre-training-based approaches (e.g., Uni-Mol \cite{NEID115}, GeoSSL \cite{NEID117}, and BindNet \cite{NEID87}), and interaction-based models (e.g., ProNet \cite{NEID121}, GET \cite{NEID123}, and CheapNet \cite{NEID2}) on the ATOM3D LBA benchmarks. Overall, the proposed method achieves the best or highly competitive performance across both the LBA30\% and LBA60\% splits. On the LBA30\% split, the proposed model achieves the highest Pearson correlation of 0.651 and Spearman correlation of 0.641. Although the RMSE on LBA30\% is marginally higher than that of CheapNet and GET, the improved correlation metrics indicate better consistency in affinity ranking across structurally diverse proteins. On the LBA60\% split, the proposed model further achieves the best overall performance. These results demonstrate strong robustness and cross-protein generalization ability under dissimilar protein scenarios.

\subsection{Effect of Graph Construction Strategies}

As defined in Eq.~\eqref{eq:dcutoff}, the intermolecular distance cutoff $d_{\text{cutoff}}$ is used in the construction of the bound complex graph for dynamic mode analysis, where it determines whether ligand and protein atoms are connected through inter-molecular interaction edges according to their spatial distance. Physically, this cutoff controls the range of protein-ligand coupling considered when deriving curvature-informed dynamic descriptors. An overly small cutoff may miss relevant local inter-molecular couplings, whereas an overly large cutoff may introduce redundant or noisy long-range connections. To investigate its influence, we evaluate the model under different $d_{\text{cutoff}}$ values and protein representation configurations, as shown in Fig.~\ref{fig:dcutoff}. The whole protein representation uses the complete protein structure provided in the dataset, while the pocket representation retains entire protein residues selected within $\alpha_{\text{dis}}=10\,\text{\AA}$ of the ligand. Overall, $d_{\text{cutoff}}=3\,\text{\AA}$ achieves the best performance across the benchmark test sets, suggesting that an appropriate cutoff effectively captures key local protein-ligand coupling patterns while avoiding excessive noisy connections. In addition, using the $10\,\text{\AA}$ binding pocket consistently outperforms or matches the whole protein representation. This observation indicates that incorporating the entire protein does not necessarily improve affinity prediction. These results demonstrate that focusing on the local binding environment provides more informative dynamic and structural patterns for the model.

\vspace{3em}
\section*{References}

\putbib[myrefs]
\end{bibunit}

\end{document}